\pdfoutput=1
\documentclass[11pt]{article}

\usepackage[table]{xcolor}
\usepackage[preprint]{acl}

\usepackage{times}
\usepackage{latexsym}

\usepackage[T1]{fontenc}
\usepackage[utf8]{inputenc}

\usepackage{microtype}

\usepackage{inconsolata}

\usepackage{graphicx}

\usepackage{tabularx}
\usepackage{lipsum} % for placeholder text
\usepackage{geometry}
\usepackage{multicol}

\title{\textsc{ReasoningFlow}: Semantic Structure of Complex Reasoning Traces}

\author{Jinu Lee, Sagnik Mukherjee, Dilek Hakkani-Tur, Julia Hockenmaier \\
  University of Illinois Urbana-Champaign \\
  \texttt{\{jinulee2, sagnikm3, dilek, juliahmr\}@illinois.edu}}

\begin{document}
\maketitle
\begin{abstract}
Large reasoning models (LRMs) generate complex reasoning traces with planning, reflection, verification, and backtracking. In this work, we introduce \textsc{ReasoningFlow}, a unified schema for analyzing the semantic structures of these complex traces. \textsc{ReasoningFlow} parses traces into directed acyclic graphs, enabling the characterization of distinct reasoning patterns as subgraph structures. This human-interpretable representation offers promising applications in understanding, evaluating, and enhancing the reasoning processes of LRMs.

\end{abstract}

\section{Introduction}

Large language models (LLMs) have shown remarkable step-by-step reasoning ability \citep{wei2022chain, kojima2022large}. In step-by-step reasoning, LLMs first generate \textit{traces} of intermediate steps that lead to the final answer instead of directly outputting the answer.

Recently, LLMs specifically trained for strong reasoning ability (Large Reasoning Models (LRMs); \textit{e.g.}, DeepSeek-R1 \citep{guo2025deepseek}, QwQ \citep{team2024qwq}) have shown outstanding performance in challenging tasks such as olympiad-level math. These models generate long reasoning traces with complex structures such as planning, verification, and backtracking \citep{gandhi2025cognitive}. While existing works analyzing the LRM-specific reasoning patterns adopt superficial methods such as keyword match \citep{yeo2025demystifying} or zero-shot classification \citep{gandhi2025cognitive}, they fail to address the deep semantic structures of the trace.

\begin{figure}[h]
    \centering
    \includegraphics[width=\linewidth]{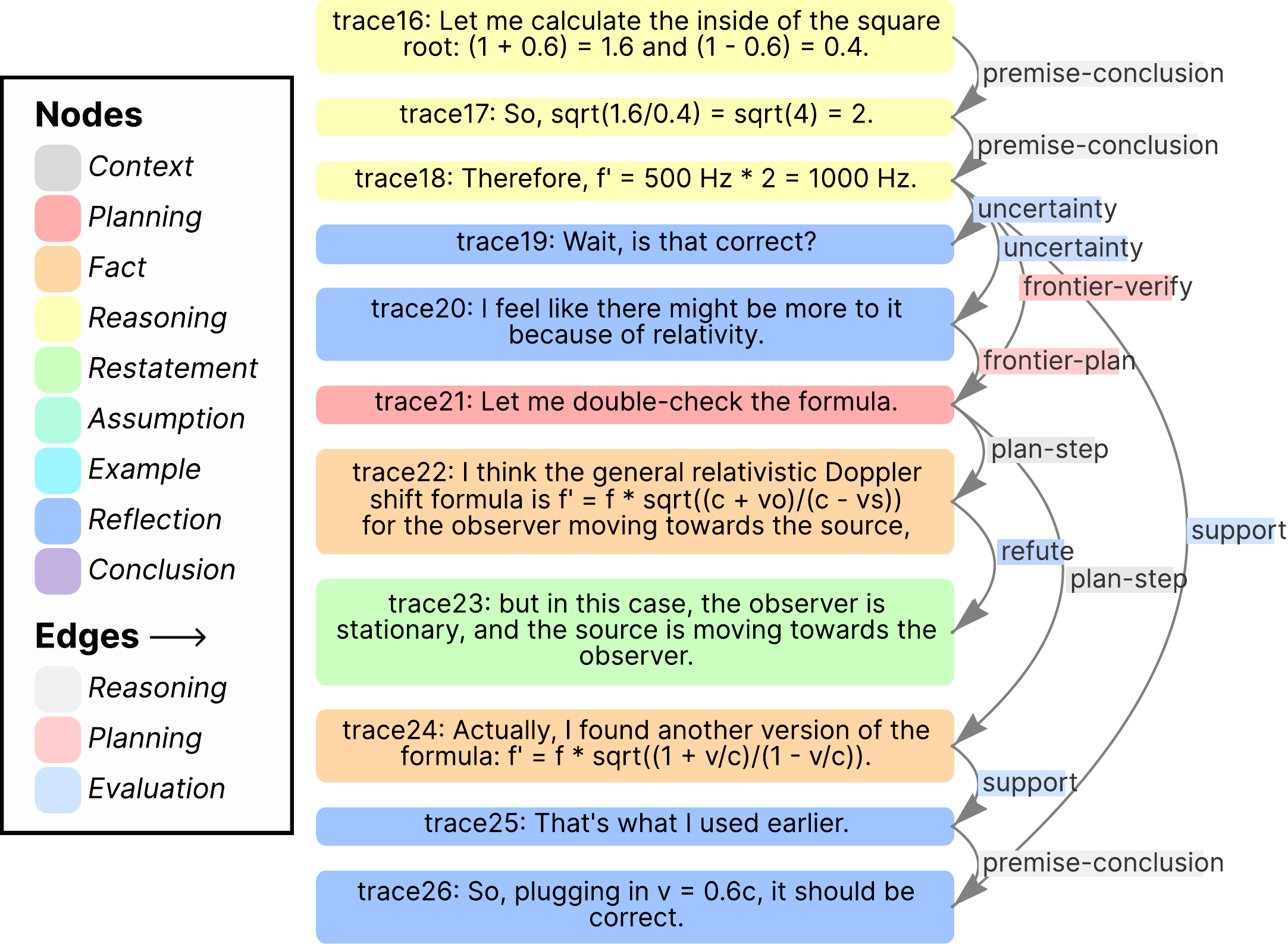}
    \caption{\textsc{ReasoningFlow} annotates the semantic structure of a complex reasoning trace. Graph-based structure provides rich contextual information about complex cognitive processes like self-verification, as shown in the example.}
    \label{fig:intro}
\end{figure}

In this work, we propose \textsc{ReasoningFlow}, a fine-grained scheme for annotating the semantic structure of LRMs' traces. By building an explicit directed acyclic graph structure, \textsc{ReasoningFlow} can characterize diverse reasoning patterns as corresponding subgraph patterns. \textsc{ReasoningFlow} enables deeper understanding of LRM-generated traces, possibly leading to explainable methods for trace validity evaluation and increasing reasoning efficiency.
% Verification vs self-verification: In the perspective of nodes, one node is verifying another. In the perspective of models, the model is self-verifying. In this paper, I chose verification (1) to emphasize the relations between nodes, and (2) to save some space!
\section{Related works}

% \subsection{Behaviors of LRMs}

% LRMs are known to demonstrate cognitive behaviors significantly distinguishable from standard instruction-tuned models, including verification (also referred to as \textit{aha-moment}), planning, and backtracking. \citep{guo2025deepseek, gandhi2025cognitive}.

% However, no attempts have been made to annotate the \textit{argumentation structure} of LRM traces guided by human intuition. \citet{yeo2025demystifying} and \citet{gandhi2025cognitive} find specific phrases that trigger behaviors like verification using lexical matching or LLM-based zero-shot classification. However, these works fail to address the structural context. 
% For instance, we observed that nodes starting with '\textit{Wait,}' are not necessarily verifying previous steps, which was assumed by (Appendix \ref{sec:appendix-stats}).

% \subsection{Semantic structures of reasoning traces}

% Previous works have attempted to analyze the structure of reasoning traces by converting them into \textit{entailment graphs} \citep{ling2023deductive, mukherjee2025premise}. In these works, the edges connect premises from the previous steps to the conclusion, forming a directed acyclic graph (DAG). However, previous works did not model complex behaviors other than entailment (\textit{e.g.}, planning and verification) observed in recent LRMs. Furtheremore, these works solely depend on LLMs in constructing the graph without human intervention, lacking parsing accuracy and explainability.

Reasoning traces of LLMs and human-written argumentative texts are both sequences of steps that lead to a specific conclusion \citep{liu2023argument}. Therefore, it motivates investigating the logical \textit{structure} of LLM-generated traces, as argumentation structures have enabled explainable evaluation \citep{stab2017parsing} and compression \citep{xu2020using} in argumentative texts.

One common approach to analyze the logical structure of traces is to use \textit{entailment graphs}, where premise steps are linked to respective conclusions to form a directed acyclic graph (DAG) \citep{ling2023deductive, mukherjee2025premise}. However, these works solely relied on LLMs for annotation without any human supervision, and their schema cannot extend to diverse reasoning patterns beyond the basic premise-conclusion relationship.
% \sagnik{major drawback was that it was not fine grained either}

Recent works on LRMs identify LRM-specific reasoning patterns by lexical match \citep{yeo2025demystifying} or LLM-based classification \citep{gandhi2025cognitive}. However, these classification-based approaches cannot provide further information about how statements relate (\textit{e.g.}, \textit{Which previous statement is being verified?}), ignoring the structural aspects of reasoning.

% For instance, we show that nodes starting with '\textit{Wait,}' are not necessarily verification or backtracking, which was assumed by \citet{yeo2025demystifying, muennighoff2025s1} (Appendix \ref{sec:appendix-stats}).
\section{\textsc{ReasoningFlow}}

\subsection{Annotation scheme}

In \textsc{ReasoningFlow}, each question and reasoning trace is segmented into \textbf{nodes}, which are contiguous, non-overlapping text snippets ranging from clauses to sentences. We define 8 node classes based on their semantic roles in the reasoning, such as Plan, Reasoning, and Reflection (Table \ref{tab:node-labels}).

\textbf{Edges} then connect these nodes to form a labeled DAG, which captures the left-to-right information flow in autoregressive generation \citep{ling2023deductive}. \textsc{ReasoningFlow} has three main edge types -- reasoning, planning, and evaluation -- with a total of 14 fine-grained labels (Table \ref{tab:edge-labels}).

\begin{figure}
    \centering
    \includegraphics[width=\linewidth]{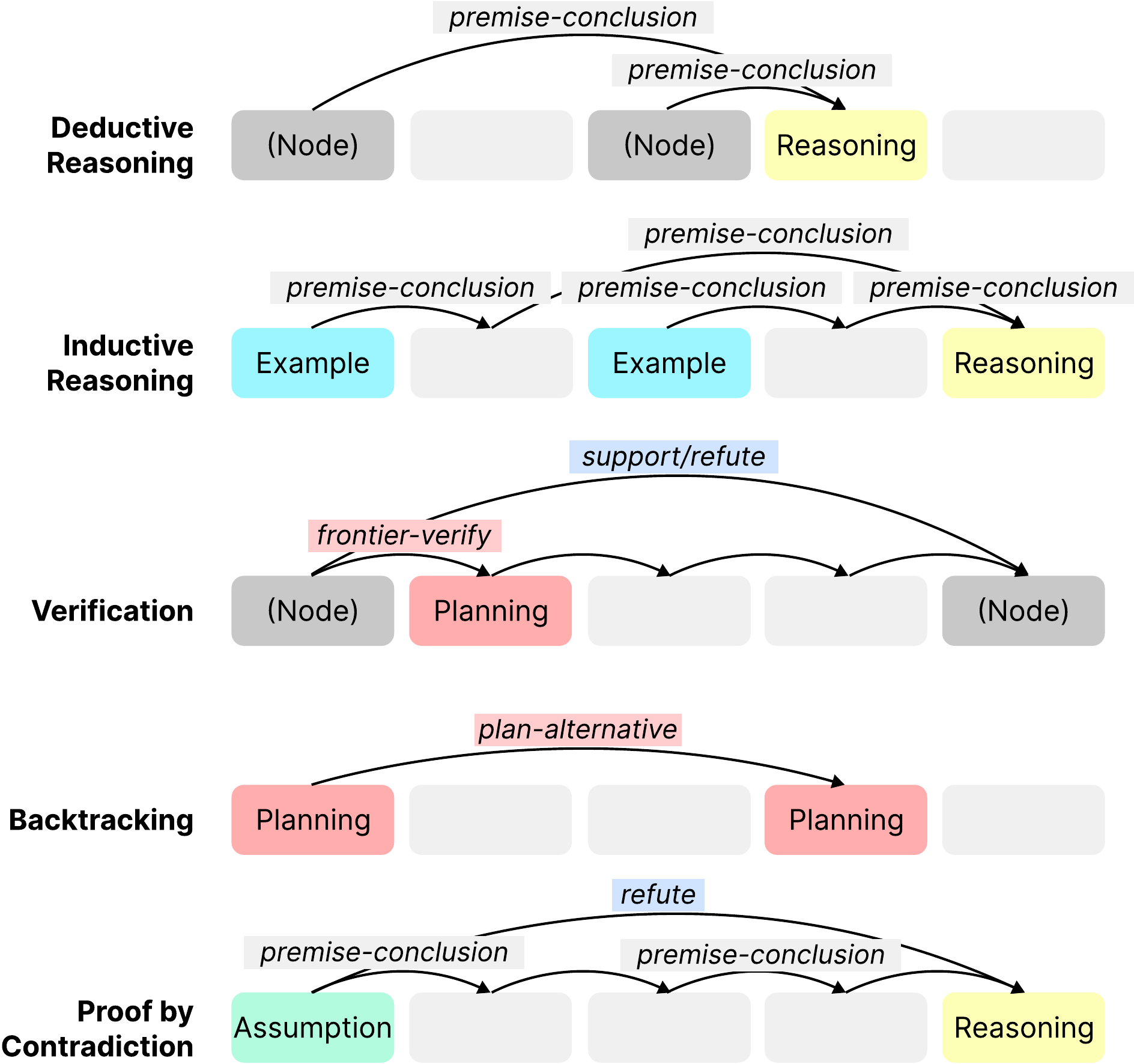}
    \caption{Examples of different reasoning patterns shown in \textsc{ReasoningFlow} graphs.}
    \label{fig:reasoning-patterns}
\end{figure}

\textsc{ReasoningFlow} reveals important reasoning structures, including deductive/inductive reasoning, verification, backtracking, and proof-by-contradiction, as subgraph patterns (Figure \ref{fig:reasoning-patterns}; Appendix \ref{appendix:subgraph-matching}). Unified graph schema of \textsc{ReasoningFlow} provides richer structural information than existing shallow methods \citep{yeo2025demystifying, gandhi2025cognitive}.

Detailed descriptions about node and edge labels can be found in Appendix \ref{sec:appendix-annotation-guide}.

\subsection{Dataset}

To construct sample annotations for \textsc{ReasoningFlow}, we sample 30 problems (15 math, 8 chemistry, 7 physics) and QwQ-32B-Preview \citep{team2024qwq} traces from Sky-T1-data \citep{sky_t1_2025}. The authors manually annotated the entire DAG, from node segmentation, classification, and edge connection using a dedicated web-based interface. Detailed statistics of \textsc{ReasoningFlow} can be found in Appendix \ref{sec:appendix-stats}.
\section{Potential applications}

\subsection{Validity evaluation}

Existing methods for evaluating LLMs' reasoning traces (\textit{e.g.}, Process Reward Models \citep{lightman2023let}, LLM-as-a-judge \citep{gu2024survey}) do not generalize to LRM traces due to their verbosity and complex structures \citep{lee2025evaluating}. When evaluating the validity of a step, edge annotations allow using only \textit{connected nodes} instead of the entire trace \citep{mukherjee2025premise}. Furthermore, \textsc{ReasoningFlow}'s node annotations allow bypassing non-reasoning steps that PRMs often falsely regard as errors (\textit{e.g.}, planning, reflection) \citep{kim2025scaling}.

% In \textsc{ReasoningFlow}, Node annotations enable tailored evaluation for different types of nodes, \textit{e.g.}, PRMs for reasoning nodes, retrieval-augmented fact checkers for facts, and entailment checking models for restatements. Furthermore, semantic structure allow \textit{scopes}. Scopes are statements that depend on certain assumptions, which are often deliberately false (\textit{e.g.}, proof by contradiction) and should be considered separately.

% Finally, \textsc{ReasoningFlow} can contribute for more efficient evaluation. LLM-as-a-judge evaluation on reasoning traces is often very costly, as it requires the full reasoning trace as an input \citep{kim2025scaling}. Using only the relevant segment of the reasoning trace based on the entailment graph is more efficient with competent accuracy \citep{ling2023deductive, mukherjee2025premise}, which is a direct consequence of \textsc{ReasoningFlow}.

\subsection{Increasing efficiency of LRMs}

LRMs are often criticized for generating overly lengthy reasoning traces for a relatively simple problem \citep{sui2025stop}. As \textsc{ReasoningFlow} can determine which nodes are necessary for deriving the final answer by analyzing the ancestors in the DAG, it can be used to compress long traces for distillation \citep{qu2025survey} or detect unpromising directions during the inference \citep{sui2025stop}.
\section{Conclusion}

In this paper, we introduce \textsc{ReasoningFlow}, an annotation scheme that can annotate complex semantic structures of reasoning traces generated by LRMs. \textsc{ReasoningFlow} annotates different semantic roles of nodes such as Reasoning, Planning, Facts, and Restatements, and connects them into a labeled directed acyclic graph that denotes interactions such as deductive reasoning, verification, and backtracking. \textsc{ReasoningFlow} can be potentially adopted to understand and improve LRMs' behaviors, including precise evaluation and increasing efficiency.

\bibliography{custom}

\begin{thebibliography}{23}
\providecommand{\natexlab}[1]{#1}

\bibitem[{Gandhi et~al.(2025)Gandhi, Chakravarthy, Singh, Lile, and Goodman}]{gandhi2025cognitive}
Kanishk Gandhi, Ayush Chakravarthy, Anikait Singh, Nathan Lile, and Noah~D Goodman. 2025.
\newblock Cognitive behaviors that enable self-improving reasoners, or, four habits of highly effective stars.
\newblock \emph{arXiv preprint arXiv:2503.01307}.

\bibitem[{Gebser et~al.(2014)Gebser, Kaminski, Kaufmann, and Schaub}]{gebser2014clingo}
Martin Gebser, Roland Kaminski, Benjamin Kaufmann, and Torsten Schaub. 2014.
\newblock Clingo= asp+ control: Preliminary report.
\newblock \emph{arXiv preprint arXiv:1405.3694}.

\bibitem[{Gu et~al.(2024)Gu, Jiang, Shi, Tan, Zhai, Xu, Li, Shen, Ma, Liu et~al.}]{gu2024survey}
Jiawei Gu, Xuhui Jiang, Zhichao Shi, Hexiang Tan, Xuehao Zhai, Chengjin Xu, Wei Li, Yinghan Shen, Shengjie Ma, Honghao Liu, and 1 others. 2024.
\newblock A survey on llm-as-a-judge.
\newblock \emph{arXiv preprint arXiv:2411.15594}.

\bibitem[{Guo et~al.(2025)Guo, Yang, Zhang, Song, Zhang, Xu, Zhu, Ma, Wang, Bi et~al.}]{guo2025deepseek}
Daya Guo, Dejian Yang, Haowei Zhang, Junxiao Song, Ruoyu Zhang, Runxin Xu, Qihao Zhu, Shirong Ma, Peiyi Wang, Xiao Bi, and 1 others. 2025.
\newblock Deepseek-r1: Incentivizing reasoning capability in llms via reinforcement learning.
\newblock \emph{arXiv preprint arXiv:2501.12948}.

\bibitem[{Kim et~al.(2025)Kim, Wu, Lee, Yue, Lee, Moon, Gashteovski, Lawrence, Hockenmaier, Neubig et~al.}]{kim2025scaling}
Seungone Kim, Ian Wu, Jinu Lee, Xiang Yue, Seongyun Lee, Mingyeong Moon, Kiril Gashteovski, Carolin Lawrence, Julia Hockenmaier, Graham Neubig, and 1 others. 2025.
\newblock Scaling evaluation-time compute with reasoning models as process evaluators.
\newblock \emph{arXiv preprint arXiv:2503.19877}.

\bibitem[{Kojima et~al.(2022)Kojima, Gu, Reid, Matsuo, and Iwasawa}]{kojima2022large}
Takeshi Kojima, Shixiang~Shane Gu, Machel Reid, Yutaka Matsuo, and Yusuke Iwasawa. 2022.
\newblock Large language models are zero-shot reasoners.
\newblock \emph{Advances in neural information processing systems}, 35:22199--22213.

\bibitem[{Lee and Hockenmaier(2025)}]{lee2025evaluating}
Jinu Lee and Julia Hockenmaier. 2025.
\newblock Evaluating step-by-step reasoning traces: A survey.
\newblock \emph{arXiv preprint arXiv:2502.12289}.

\bibitem[{Lightman et~al.(2023)Lightman, Kosaraju, Burda, Edwards, Baker, Lee, Leike, Schulman, Sutskever, and Cobbe}]{lightman2023let}
Hunter Lightman, Vineet Kosaraju, Yuri Burda, Harrison Edwards, Bowen Baker, Teddy Lee, Jan Leike, John Schulman, Ilya Sutskever, and Karl Cobbe. 2023.
\newblock Let's verify step by step.
\newblock In \emph{The Twelfth International Conference on Learning Representations}.

\bibitem[{Ling et~al.(2023)Ling, Fang, Li, Huang, Lee, Memisevic, and Su}]{ling2023deductive}
Zhan Ling, Yunhao Fang, Xuanlin Li, Zhiao Huang, Mingu Lee, Roland Memisevic, and Hao Su. 2023.
\newblock Deductive verification of chain-of-thought reasoning.
\newblock \emph{Advances in Neural Information Processing Systems}, 36:36407--36433.

\bibitem[{Liu et~al.(2023)Liu, Schlegel, Batista-Navarro, and Ananiadou}]{liu2023argument}
Boyang Liu, Viktor Schlegel, Riza~Theresa Batista-Navarro, and Sophia Ananiadou. 2023.
\newblock Argument mining as a multi-hop generative machine reading comprehension task.
\newblock In \emph{Findings of the Association for Computational Linguistics: EMNLP 2023}, pages 10846--10858.

\bibitem[{Mann and Thompson(1987)}]{mann1987rhetorical}
William~C Mann and Sandra~A Thompson. 1987.
\newblock \emph{Rhetorical structure theory: A theory of text organization}.
\newblock University of Southern California, Information Sciences Institute Los Angeles.

\bibitem[{Marjanović et~al.(2025)Marjanović, Patel, Adlakha, Aghajohari, BehnamGhader, Bhatia, Khandelwal, Kraft, Krojer, Lù, Meade, Shin, Kazemnejad, Kamath, Mosbach, Stańczak, and Reddy}]{marjanovic2025deepseekr1thoughtologyletsthink}
Sara~Vera Marjanović, Arkil Patel, Vaibhav Adlakha, Milad Aghajohari, Parishad BehnamGhader, Mehar Bhatia, Aditi Khandelwal, Austin Kraft, Benno Krojer, Xing~Han Lù, Nicholas Meade, Dongchan Shin, Amirhossein Kazemnejad, Gaurav Kamath, Marius Mosbach, Karolina Stańczak, and Siva Reddy. 2025.
\newblock \href {https://arxiv.org/abs/2504.07128} {Deepseek-r1 thoughtology: Let's <think> about llm reasoning}.
\newblock \emph{Preprint}, arXiv:2504.07128.

\bibitem[{Mukherjee et~al.(2025)Mukherjee, Chinta, Kim, Sharma, and Hakkani-T{\"u}r}]{mukherjee2025premise}
Sagnik Mukherjee, Abhinav Chinta, Takyoung Kim, Tarun~Anoop Sharma, and Dilek Hakkani-T{\"u}r. 2025.
\newblock Premise-augmented reasoning chains improve error identification in math reasoning with llms.
\newblock \emph{arXiv preprint arXiv:2502.02362}.

\bibitem[{NovaSky(2025)}]{sky_t1_2025}
NovaSky. 2025.
\newblock \href {https://novasky-ai.github.io/posts/sky-t1} {Sky-t1: Train your own o1 preview model within \$450}.
\newblock Accessed: 2025-01-09.

\bibitem[{Qu et~al.(2025)Qu, Li, Su, Sun, Yan, Liu, Cui, Liu, Liang, He et~al.}]{qu2025survey}
Xiaoye Qu, Yafu Li, Zhaochen Su, Weigao Sun, Jianhao Yan, Dongrui Liu, Ganqu Cui, Daizong Liu, Shuxian Liang, Junxian He, and 1 others. 2025.
\newblock A survey of efficient reasoning for large reasoning models: Language, multimodality, and beyond.
\newblock \emph{arXiv preprint arXiv:2503.21614}.

\bibitem[{Qwen(2024)}]{team2024qwq}
Qwen. 2024.
\newblock Qwq: Reflect deeply on the boundaries of the unknown.
\newblock \emph{HuggingFace}.

\bibitem[{Stab and Gurevych(2017)}]{stab2017parsing}
Christian Stab and Iryna Gurevych. 2017.
\newblock Parsing argumentation structures in persuasive essays.
\newblock \emph{Computational Linguistics}, 43(3):619--659.

\bibitem[{Sui et~al.(2025)Sui, Chuang, Wang, Zhang, Zhang, Yuan, Liu, Wen, Chen, Hu et~al.}]{sui2025stop}
Yang Sui, Yu-Neng Chuang, Guanchu Wang, Jiamu Zhang, Tianyi Zhang, Jiayi Yuan, Hongyi Liu, Andrew Wen, Hanjie Chen, Xia Hu, and 1 others. 2025.
\newblock Stop overthinking: A survey on efficient reasoning for large language models.
\newblock \emph{arXiv preprint arXiv:2503.16419}.

\bibitem[{Thorne et~al.(2018)Thorne, Vlachos, Christodoulopoulos, and Mittal}]{thorne2018fever}
James Thorne, Andreas Vlachos, Christos Christodoulopoulos, and Arpit Mittal. 2018.
\newblock Fever: a large-scale dataset for fact extraction and verification.
\newblock In \emph{Proceedings of the 2018 Conference of the North American Chapter of the Association for Computational Linguistics: Human Language Technologies, Volume 1 (Long Papers)}, pages 809--819.

\bibitem[{Wang et~al.(2025)Wang, Liu, Xu, Liang, Chen, He, Song, Yu, Li, Zhang et~al.}]{wang2025thoughts}
Yue Wang, Qiuzhi Liu, Jiahao Xu, Tian Liang, Xingyu Chen, Zhiwei He, Linfeng Song, Dian Yu, Juntao Li, Zhuosheng Zhang, and 1 others. 2025.
\newblock Thoughts are all over the place: On the underthinking of o1-like llms.
\newblock \emph{arXiv preprint arXiv:2501.18585}.

\bibitem[{Wei et~al.(2022)Wei, Wang, Schuurmans, Bosma, Xia, Chi, Le, Zhou et~al.}]{wei2022chain}
Jason Wei, Xuezhi Wang, Dale Schuurmans, Maarten Bosma, Fei Xia, Ed~Chi, Quoc~V Le, Denny Zhou, and 1 others. 2022.
\newblock Chain-of-thought prompting elicits reasoning in large language models.
\newblock \emph{Advances in neural information processing systems}, 35:24824--24837.

\bibitem[{Xu et~al.(2020)Xu, {\v{S}}avelka, and Ashley}]{xu2020using}
Huihui Xu, Jarom{\'\i}r {\v{S}}avelka, and Kevin~D Ashley. 2020.
\newblock Using argument mining for legal text summarization.
\newblock In \emph{Legal Knowledge and Information Systems}, pages 184--193. IOS Press.

\bibitem[{Yeo et~al.(2025)Yeo, Tong, Niu, Neubig, and Yue}]{yeo2025demystifying}
Edward Yeo, Yuxuan Tong, Morry Niu, Graham Neubig, and Xiang Yue. 2025.
\newblock Demystifying long chain-of-thought reasoning in llms.
\newblock \emph{arXiv preprint arXiv:2502.03373}.

\end{thebibliography}

\newpage

\appendix
\section{\textsc{ReasoningFlow} scheme}
\label{sec:appendix-annotation-guide}

The objective of \textsc{ReasoningFlow} is to identify the logical structure of Chain-of-thought reasoning traces, especially ones generated by large reasoning models. \textsc{ReasoningFlow} has two main structural components, namely \textbf{nodes} and \textbf{edges}.

\paragraph{Nodes} Reasoning traces are segmented into nodes, which are syntactically non-overlapping segments from clauses to multi-line paragraphs. The authors manually determined the node boundary based on semantic atomicity. Basic segmentation units are sentences or lines. However, if the sentence contains clauses that might be assigned different node labels, they are separated. For instance, the sentence \textit{"I should also consider the possibility of forming 1-butene or other isomers\textcolor{red}{\textbf{,}} but given the structure, 2-methylbut-1-ene seems to be the only plausible product."} can be segmented by the red comma, where the first segment is a Planning node and the second is a Reasoning node.

Each node is assigned a unique label based on its semantic role. The detailed definitions of the node labels are presented in Table \ref{tab:node-labels}.

\begin{table*}[ht]
\centering \small
\begin{tabularx}{\textwidth}{clX}
\hline
& \textbf{Node} & \textbf{Description} \\
\hline
\cellcolor[HTML]{D9D9D9} & Context & Context corresponds to text provided to the model before the first generated token. It typically includes questions and retrieved documents, if there are any. \\ \hline
\cellcolor[HTML]{FFADAD} & Planning & Planning introduces the content of following nodes. It might be coarse, high-level directions that affect tens of nodes (\textit{Let's try to simplify the equation.}) to highly specific, local plans that affect only one or two nodes (\textit{Let's add 2 to both sides.}). The rule of thumb is to check if (1) a node includes active verbs like find, recall, check, solve, replace with first-person subjects, and (2) the following node is reasoning (fact, restatement, assumption, example) with contents highly related to the current node.
\begin{itemize}
\setlength\itemsep{0pt}
    \item Introducing overall directions for multiple upcoming nodes
    \item Fine-grained plan specifically for the next step
    \item Phrases that initiate the verification process (\textit{Let's double-check this.})
\end{itemize}

\\ \hline
\cellcolor[HTML]{FFD6A5} & Fact & Fact refers to phrases that contain general external knowledge, (1) not specific to this problem, (2) not given in context, (3) not deduced from other facts. The validity of the Fact node can be judged by external search, as in the fact verification task \citep{thorne2018fever}. An intuitive way to determine Fact is to see (1) if the claim can be directly entailed or contradicted from a Wikipedia segment, and (2) if it is not logically deduced from other traces.
\begin{itemize}
\setlength\itemsep{0pt}
    \item Theorems, laws, and rules
    \item Trivia, commonsense, and world knowledge
    \item Values of well-known constants or measurements
    \item Non-specific facts (\textit{e.g.}, I recall that there is an equation for A)
\end{itemize}
\\ \hline
\cellcolor[HTML]{FDFFB6} & Reasoning & Reasoning includes any sort of deductive/inductive/abductive inference. 
\begin{itemize}
\setlength\itemsep{0pt}
    \item Applying facts (numbers) to rules (equations)
    \item Simplifying equations or logical statements
    \item Inductive reasoning based on examples
    \item Meta-reasoning on facts ("\textit{This fact is not applicable in our case due to ...}"). If meta-reasoning is about its previous reasoning steps, it should be classified as Reflection.
\end{itemize}
\\ \hline
\cellcolor[HTML]{CAFFBF} & Restatement & Restatement is when the model rephrases the preceding text (context/previously presented steps). Typically, Restatement includes terms like ‘as seen previously’, ‘already stated’, and ‘the problem states’. There is often a semantically equivalent node above in the trace while the context suggests that the model is re-deriving it, particularly during the verification process. These nodes should be categorized as Reasoning rather than Restatement. \\ \hline
\cellcolor[HTML]{B3FBDf} & Assumption & If one assumes a certain proposition for further reasoning process, it is an Assumption. Assumption defines a \textit{scope}, so that the statements that take the assumption as a premise should be evaluated accordingly.
\begin{itemize}
\setlength\itemsep{0pt}
    \item Assuming an underspecified condition based on common practice (\textit{e.g.}, assuming standard temperature/pressure (STP) condition in chemistry problems)
    \item Branching, where all possible cases are divided into finite possibilities. It is distinguished from Example nodes, as branching involves deductive reasoning (\textit{by considering all possibilities, $A, B, C$, we can deduce that...}) while examples are related to inductive reasoning (\textit{Given the pattern observed from examples, we can conclude that...}).
    \item Proof by contradiction.
\end{itemize}
\\ \hline
\cellcolor[HTML]{9BF6FF} & Example & Example is when a general concept/condition is provided first, then followed by a specific example of such concept/condition. Typical use cases of Example are for pattern finding, which often leads to inductive reasoning. \\ \hline
\cellcolor[HTML]{A0C4FF} & Reflection & This includes clauses that express confidence, feelings, impressions, and doubt (questioning without intent to correct). \\ \hline
\cellcolor[HTML]{C3B1E1} & Conclusion & This node includes the model's final answer. There should be only one consecutive set of nodes that have this label. If the question requires simple answers, such as in multiple-choice questions or short-answer questions, only one node from the entire trace can be assigned this label. \\
\end{tabularx}
\caption{Node labels of \textsc{ReasoningFlow}.}
\label{tab:node-labels}
\normalsize
\end{table*}

\paragraph{Edges} Edges show the semantic relationship between different nodes. \textsc{ReasoningFlow} assumes that the edges always start from the earlier nodes (predecessors) and end at the later nodes (successors), forming a left-to-right directed acyclic graph (DAG). While transitive, constituent trees are the standard structure in discourse/argumentation parsing \citep{mann1987rhetorical, stab2017parsing}, DAG acknowledges the left-to-right information flow created by the autoregressive generation of LLM and widely accepted in reasoning trace structure analysis literature \citep{ling2023deductive, mukherjee2025premise, marjanovic2025deepseekr1thoughtologyletsthink}. Another benefit of DAG is that it allows a simple 2-stage parsing algorithm: (1) select the direct predecessors as a binary classification problem (\textit{Is there an edge between two nodes?}) and (2) label the edges.

There are two basic axioms for drawing an edge between two nodes. First, the set of predecessors should include antecedents for all the pronouns and definitions for all the symbols used in the successor node. This ensures that one can fully understand and evaluate the successor by only looking at the connected predecessors, a deliberate design choice to support graph-based evaluation \citep{ling2023deductive, mukherjee2025premise}. Second, if there are multiple instances of lexically similar predecessors (due to restatements), one must first follow the coreference resolution/symbol definition (typically within the same paragraph), and otherwise choose the earliest node. This rule preserves the local semantic structure (\textit{e.g.}, paragraph level) when analyzing long reasoning traces.

There are 3 coarse edge label categories (Reasoning, Planning, and Evaluation). Each category corresponds to specific sub-structures of reasoning. First, Reasoning edges capture conventional reasoning processes (\textit{e.g.}, deductive, inductive reasoning). Next, Planning edges annotate high-level structures of the solution introduced by top-down planning \citep{gandhi2025cognitive}, and finally, Evaluation edges show the LRM's attitude (confidence, sentiment, ...) about the self-generated reasoning paths. There are 14 fine-grained edge labels under these three categories; detailed definitions can be found in Table \ref{tab:edge-labels}.

\begin{table*}[ht]
\centering \small
\begin{tabularx}{\textwidth}{clX}
\hline
& \textbf{Edge} & \textbf{Description} \\
\hline
& \multicolumn{2}{l}{\textit{Planning}} \\
\hline
\cellcolor[HTML]{FFCDCD} & Frontier-Plan & This edge connects facts/reasoning/restatements/examples/... to a plan, when the content of the previous node triggers the planning, \textit{i.e.}, serves as a premise for following nodes (connected with Plan-Step edge). Typical examples include (3) deciding high-level plans based on known facts and (2) indicating how to solve (simplify, substitute, calculate) a certain equation at a low level. \\ \hline
\cellcolor[HTML]{FFC6C6} & Frontier-Verify & This edge specifically labels the relationship between a reasoning node and a Planning node initiating the verification process. Typical subsequent nodes include phrases like "\textit{Let's verify this}", "\textit{I should check if there is any error}". \\ \hline
\cellcolor[HTML]{FFBDBD} & Plan-Subplan & This edge connects a plan and a subplan. Subplans decompose the high-level plan into small, straightforward goals. If the plans can be executed in parallel, edges should connect the high-level plan to all subplans; if the subplans are meant to execute sequentially, edges should connect to the first plan only. \\ \hline
\cellcolor[HTML]{FFB6B6} & Plan-Next plan & This edge connects multiple plans that are executed sequentially. The subsequent nodes often start with phrases like "\textit{Next}", "\textit{Second}", "\textit{Then}". \\ \hline
\cellcolor[HTML]{FFADAD} & Plan-Alternative & This edge denotes when the model is suggesting an alternative plan from the previous one. The subsequent nodes often include "\textit{Alternatively}", "\textit{Let me think in another direction}", or other similar phrases. \\ \hline
& \multicolumn{2}{l}{\textit{Reasoning}} \\
\hline
\cellcolor[HTML]{F0F0F0} & Premise-Conclusion & This relationship represents the basic premise-conclusion reasoning relationship. It can include deductive/inductive/abductive reasoning. \\ \hline
\cellcolor[HTML]{E8E8E8} & Plan-Step & This relationship represents the relationship between a plan and its implementation. This edge should connect the plan and the first nodes of parallel reasoning chains. For instance, if the plan reads "\textit{Let me check the first three examples}", the Plan-step edge should connect the plan to all three examples. \\ \hline
\cellcolor[HTML]{E0E0E0} & Concept-Example & This edge label applies to Example nodes, connecting the example to what it exemplifies. When both nodes are facts, it should be annotated as Fact-Detail. \\ \hline
\cellcolor[HTML]{D8D8D8} & Fact-Detail & This edge connects general facts to more specific facts. This edge is often used when the LLM generates consequent nodes of facts, and the following nodes describe the details of the initial fact. \\ \hline
\cellcolor[HTML]{D0D0D0} & Restatement & This edge label applies to Restatement nodes, connecting the original statement to the restatement. \\ \hline
\cellcolor[HTML]{C8C8C8} & Correction & This edge label applies when a node \textit{corrects} the previous one by reasoning, by providing an alternative conclusion. It should be distinguished from simple reflection/planning that does not correct a previous statement. This edge, alongside Refute, implies that the model judges a specific node as incorrect. \\ \hline
& \multicolumn{2}{l}{\textit{Evaluation}} \\
\hline
\cellcolor[HTML]{D0E4FF} & Support & This edge connects a node to a following node that \textit{supports} the previous one. Two typical cases include (1) Reflection statements such as "\textit{That looks good}", or (2) connecting the conclusion of the positive verification to the verification goal. \\ \hline
\cellcolor[HTML]{C8D8FF} & Refute & This edge connects one node and another that refutes it. Examples include reflection statements like "\textit{This seems wrong}", or reasoning nodes that refute a previous conclusion but do not correct it, \textit{i.e.}, "\textit{Therefore, it is wrong}". \\ \hline
\cellcolor[HTML]{C0D4FF} & Uncertainty & This edge represents all other reflection statements that do not fall into Support/Refute. Typical use cases are reflection of uncertainty ("\textit{I am not sure/certain}"), confusion ("\textit{I am confused/lost}"), lack of knowledge ("\textit{I don’t know}"), lack of confidence ("\textit{I am a bit rusty on this}"), and anomaly ("\textit{This seems weird}"). \\
\hline
\end{tabularx}
\caption{Edge labels of \textsc{ReasoningFlow}.}
\label{tab:edge-labels}
\normalsize
\end{table*}
\section{Subgraph matching}
\label{appendix:subgraph-matching}

\textsc{ReasoningFlow} reveals reasoning behaviors as specific subgraph patterns. Hence, we develop a subgraph querying engine to detect user-specified patterns based on Answer Set Programming (ASP) \citep{gebser2014clingo}. ASP provides an efficient way to query complex graphs based on logic programming syntax.

In the querying engine, \textsc{ReasoningFlow} graphs are expressed as set of predicates, \texttt{node/2} or \texttt{edge/3}\footnote{In ASP, \texttt{pred/N} indicates that the predicate \texttt{pred} has $N$ arguments (arity of $N$). For instance, the predicate of this expression \texttt{a(X, Y)} can be expressed as \texttt{a/2}.}. These predicates store the node and edge information with labels, e.g., \texttt{node(trace0, "restatement")}, \texttt{edge(trace0, trace1, "frontier-plan")}. Furthermore, auxiliary predicates are automatically added for user convenience. For instance, \texttt{connected/2} denote the connectivity of two nodes in the transitive closure, and \texttt{distance/3} tells the distance between two nodes. Using these predicates, one can query the graph structure to retrieve the tuple of node identifiers that satisfy the condition.

\begin{figure}
    \centering
    \includegraphics[width=\linewidth]{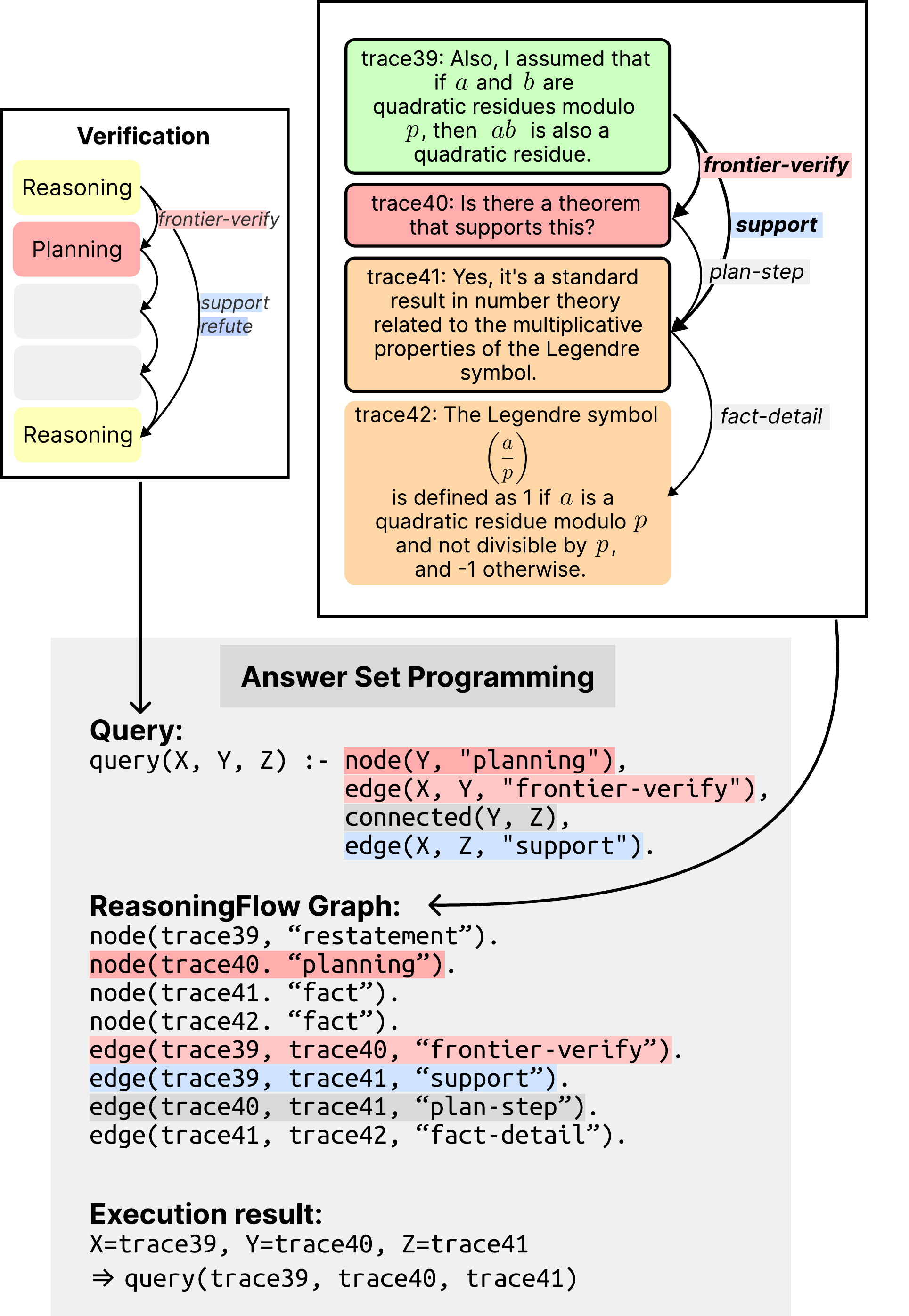}
    \caption{Example of subgraph matching. Given a specific pattern (top-left) and a \textsc{ReasoningFlow} graph (top-right), the engine executes the corresponding ASP query and finds the tuple of node identifiers (bottom). Query conditions and matching patterns are highlighted in the same color.}
    \label{fig:subgraph-matching}
\end{figure}

Figure \ref{fig:subgraph-matching} demonstrates how a specific reasoning pattern (verification) can be queried in our engine. The query searches for three nodes (\texttt{X, Y, Z}) that satisfies (1) \texttt{Y} is a planning node, (2) there exists a node \texttt{X} where \texttt{X}$\rightarrow$\texttt{Y} edge is labeled Frontier-Verify, and (3) there exists a node \texttt{Z} that is connected to \texttt{Y} under transitive closure and directly supports/refutes \texttt{X}. The engine searches the graph to find the node identifiers that satisfy all conditions, ultimately detecting the verification pattern spanning nodes \texttt{trace39, 40, 41}.

\begin{figure*}[ht]
    \centering
    \includegraphics[width=1.0\linewidth]{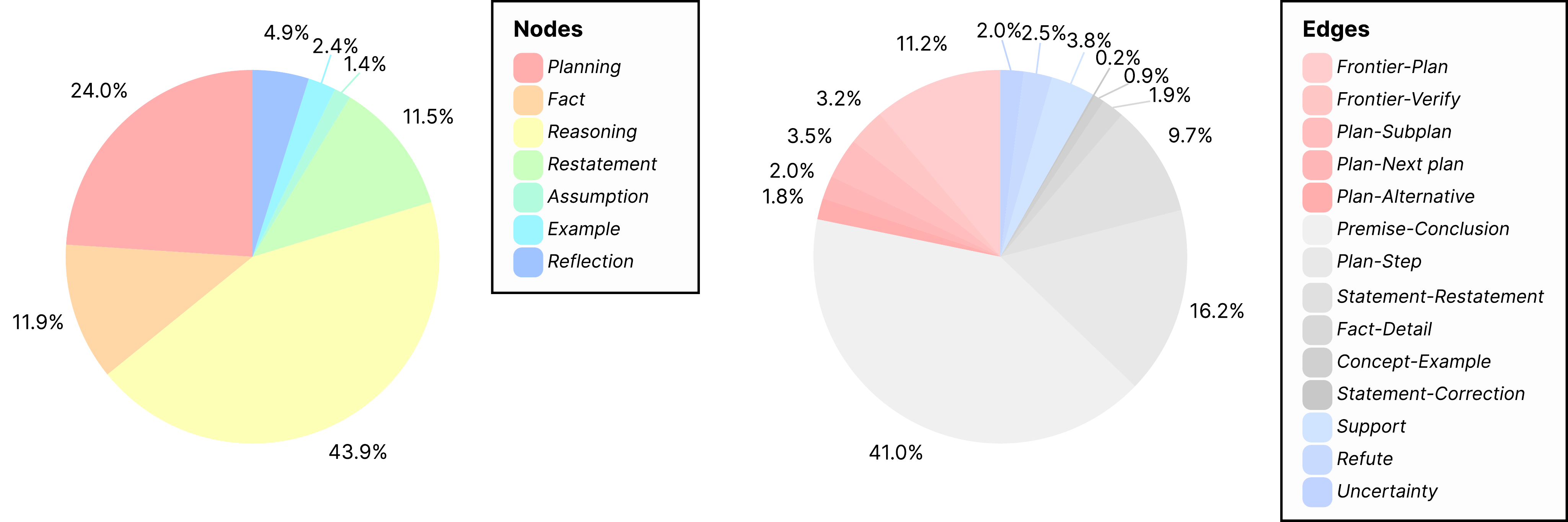}
    \caption{Edge and node label distributions of \textsc{ReasoningFlow}. Segments are ordered counterclockwise from the top, matching the item orders in the legend. In edge labels, red, gray, and blue labels fall into Planning, Reasoning, and Evaluation categories, respectively (Table \ref{tab:edge-labels}).}
    \label{fig:label-distribution}
\end{figure*}

\section{\textsc{ReasoningFlow} statistics}

\label{sec:appendix-stats}

In this section, we provide statistics of the \textsc{ReasoningFlow} dataset.

\paragraph{Response length} LRMs, including the QwQ-32B-Preview model \citep{team2024qwq} that was used to sample the responses in this study, are known to generate much longer tokens than non-reasoning LLMs \citep{wang2025thoughts}. The average number of tokens in 30 responses used in this study is 1,550.76 tokens.

Reasoning traces are further segmented into nodes, which are semantically atomic units with granularity ranging from clauses to paragraphs. Each response in \textsc{ReasoningFlow} has an average of 44.21 nodes.

\paragraph{Label distribution} 

Figure \ref{fig:label-distribution} shows the distribution of nodes and edge labels in \textsc{ReasoningFlow}. 

The most frequent node in \textsc{ReasoningFlow} is Reasoning nodes, followed by Planning, Fact, and Restatement. These nodes together make up 91.3\% of the total nodes, serving as the basic building blocks for reasoning traces. Examples and Assumptions are relatively rare, which implies that strategies like inductive reasoning or proof by contradiction are not often used in the dataset. This might have stemmed from the task distribution (math and science), where deductive reasoning is highly dominant compared to other indirect strategies.

Similarly, the edge labels exhibit a highly skewed distribution towards deductive reasoning. Premise-Conclusion takes account of 41.0\% of total edges, followed by Plan-Step and Frontier-Plan edges that often occur in low-level reasoning structures (\textit{e.g.}, simplifying an equation).
\section{Sample \textsc{ReasoningFlow} graph}
\label{sec:appendix-sample}

\begin{figure*}[p]
    \centering
    \includegraphics[width=0.85\linewidth]{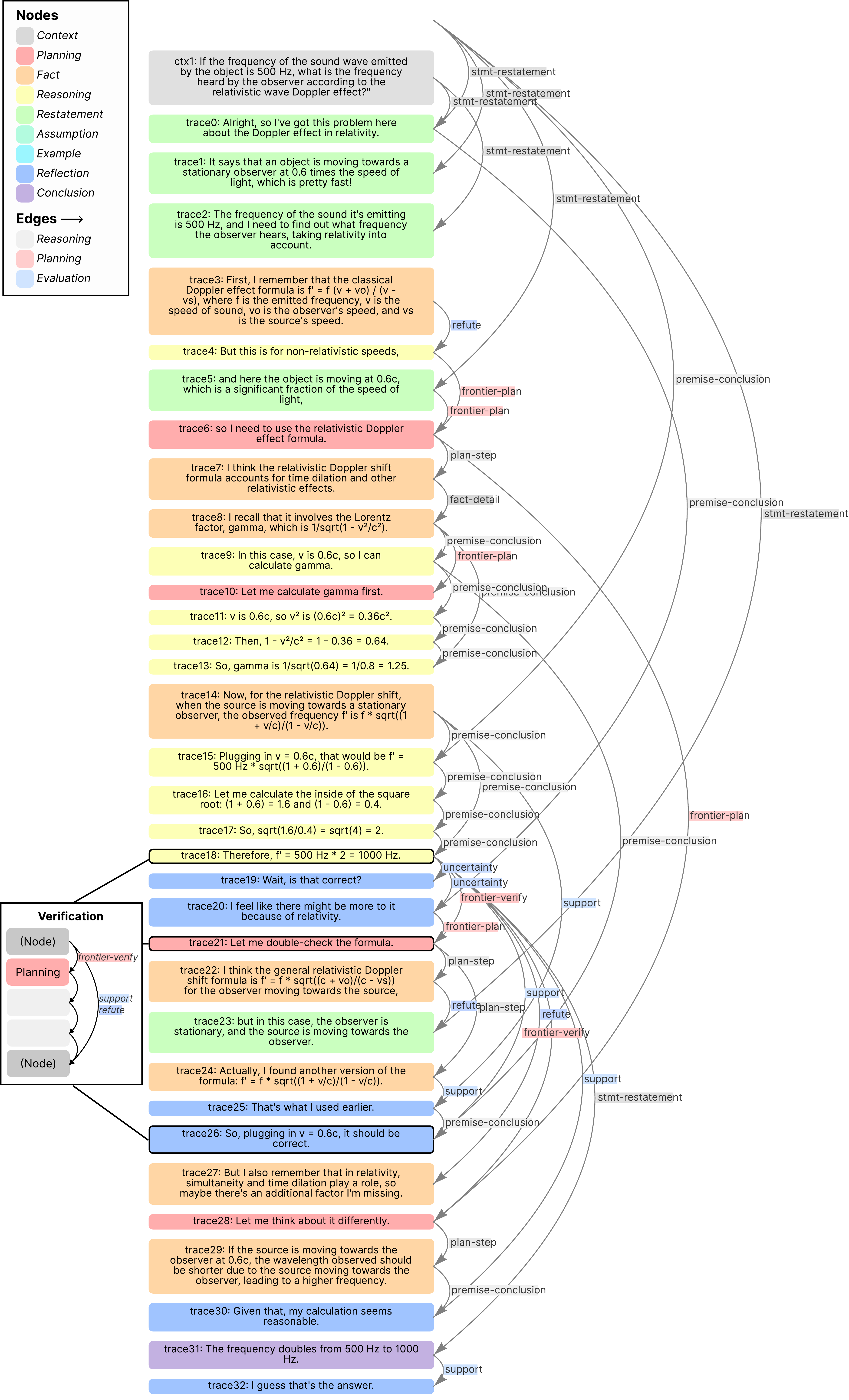}
    \caption{An example annotation of \textsc{ReasoningFlow}, constructed from a physics question (\texttt{ctx} nodes) and the response from QwQ-32B-Preview (\texttt{trace} nodes). Bordered nodes (\texttt{trace18, 21, 26}) together form a verification subgraph, which can be identified by the subgraph matching rule visualized on the left side of the figure.}
    \label{fig:full-example-1}
\end{figure*}

\begin{figure*}[p]
    \centering
    \includegraphics[width=0.6\linewidth]{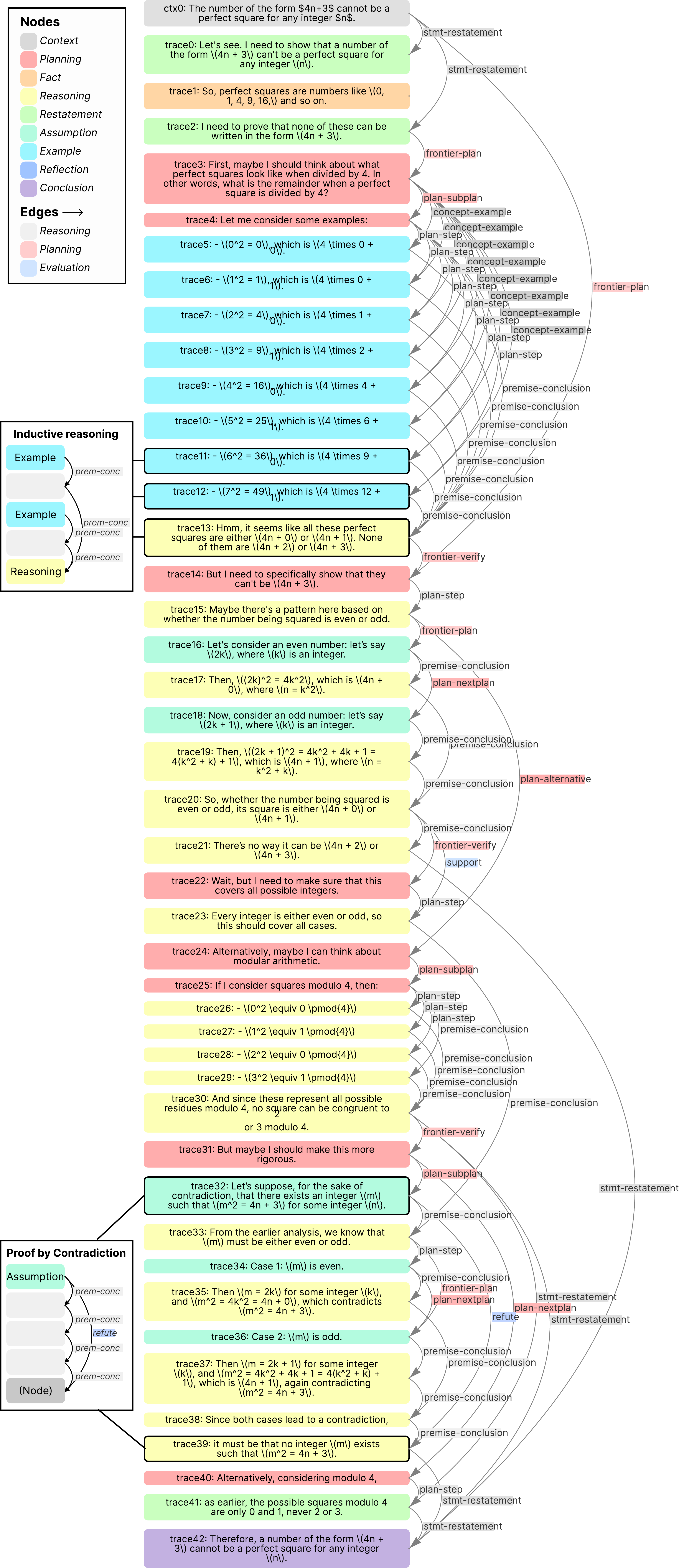}
    \caption{An example annotation of \textsc{ReasoningFlow}, constructed from a math question (\texttt{ctx} nodes) and the response from QwQ-32B-Preview (\texttt{trace} nodes). The inductive reasoning and proof by contradiction subgraphs can be identified according to the rule presented on the left side of the figure.}
    \label{fig:full-example-2}
\end{figure*}

Two sample \textsc{ReasoningFlow} graphs are shown in Figures \ref{fig:full-example-1} and \ref{fig:full-example-2}.

\end{document}